\documentclass{article}

\usepackage{arxiv}

\usepackage[utf8]{inputenc} 
\usepackage[T1]{fontenc}    
\usepackage{hyperref}       
\usepackage{url}            
\usepackage{booktabs}       
\usepackage{amsfonts}       
\usepackage{nicefrac}       
\usepackage{microtype}      
\usepackage{lipsum}
\usepackage{graphicx}
\usepackage{multirow}
\usepackage{listings}
\graphicspath{ {./images/} }

\usepackage{todonotes}

\title{Semantic Answer Type and \\Relation Prediction Task (SMART 2021)}

\author{
Nandana Mihindukulasooriya \\
IBM Research AI\\ 
  \texttt{nandana.m@ibm.com} \\
\And
  Mohnish Dubey\\
  University of Bonn \\and Fraunhofer IAIS\\
  \texttt{dubey@cs.uni-bonn.de} \\
\And
Alfio Gliozzo \\
IBM Research AI\\ 
  \texttt{gliozzo@us.ibm.com} \\
\And
Jens Lehmann \\
University of Bonn and Fraunhofer IAIS\\ 
\texttt{jens.lehmann@cs.uni-bonn.de} \\
\And
Axel-Cyrille Ngonga Ngomo \\
Universität Paderborn\\ 
\texttt{axel.ngonga@upb.de} \\
\And
Ricardo Usbeck \\
Conversational AI, Fraunhofer IAIS Dresden\\
\texttt{ricardo.usbeck@iais.fraunhofer.de} \\
\And
Gaetano Rossiello \\
IBM Research, USA\\
\texttt{gaetano.rossiello@ibm.com} \\ 
\And
Uttam Kumar \\
 University of Bonn, Germany\\
\texttt{uttam1216@gmail.com} \\
}

\begin{document}
\maketitle
\begin{abstract}
Each year the International Semantic Web Conference organises a set of Semantic Web Challenges to establish competitions that will advance state of the art solutions in some problem domains. The \textit{Semantic Answer Type and Relation Prediction Task} (SMART) task is one of the ISWC 2021 Semantic Web challenges. This is the second year of the challenge after a successful SMART 2020 at ISWC 2020. This year's version focuses on two sub-tasks that are very important to Knowledge Base Question Answering (KBQA): Answer Type Prediction and Relation Prediction. Question type and answer type prediction can play a key role in knowledge base question answering systems providing insights about the expected answer that are helpful to generate correct queries or rank the answer candidates. More concretely, given a question in natural language, the first task is, to predict the answer type using a target ontology (\textit{e.g.}, \textit{DBpedia} or \textit{Wikidata}. Similarly, the second task is to identify relations in the natural language query and link them to the relations in a target ontology. This paper discusses the task descriptions, benchmark datasets, and evaluation metrics. More information, please visit \url{https://smart-task.github.io/2021/}.

\end{abstract}

\keywords{Answer Type Prediction \and Relation Linking \and Question Answering \and ISWC Semantic Web Challenge}

\section{Background}

Question Answering (QA) is a popular task in Natural Language Processing and Information Retrieval. The goal of a QA engine is commonly to answer a natural language question based on information contained in (a potentially predefined set of) datasets. An instantiation of the QA task is reading comprehension, in which the expected answers can be either a segment of text or span from the corresponding reading passage of text. In contrast, Knowledge Graph Question Answering (KGQA), the answer is extracted from a structured knowledge graph using a query. Typically, these knowledge graphs are in RDF and the queries are formulated using the SPARQL query language. 

In KGQA, the expected answer can either be (1) a set of entities or literals from a given knowledge graph $G$ or (2) derived from an aggregation of facts found in $G$. Many available datasets for the KGQA task originate from the ISWC community (e.g., QALD~\cite{ngomo20189th}, LC-QuAD 1.0~\cite{trivedi2017lc}, LC-QuAD 2.0~\cite{dubey2019lc}, SimpleQuestions-DBpedia~\cite{azmy2018farewell}, SimpleQuestions-Wikidata~\cite{wikidata-benchmark}).

In KGQA, given a natural language query, the final goal is to formulate a logical query that can generate the answer to the questions. In SMART 2021, we will focus on two sub-tasks that are relevant to KBQA: (a) answer type prediction, and (b) relation prediction.  SMART 2021 is the second edition of the SMART task challenge (\url{https://smart-task.github.io/}). Its first incarnation happened at ISWC 2020 with 11 submitting systems in total, advancing the state of the art of answer type classification. The proceedings of the first edition is published in CEUR~\cite{DBLP:conf/semweb/2020smart} which consists of 8 submissions~\cite{perevalov2020augmentation, setty2021semantic, nikas2020two, steinmetz2020coala, ammar2020methodology, kertkeidkachorn2020hierarchical, vallurupalli2020fine, bill2020question}.

Classifying a question's answer type plays a key role in KGQA~\cite{harabagiu2000falcon,allam2012question}. The questions themselves can be generally classified based on Wh-terms (\textit{Who}, \textit{What}, \textit{When}, \textit{Where}, \textit{Which}, \textit{Whom}, \textit{Whose}, \textit{Why}). The answer type classification is the task of determining the type of the expected answer based on the query.  In literature, such answer type classifications is performed as a short-text classification task using a set of coarse-grained types, with either 6 types~\cite{zhao2015self,zhou2015c}
or fine-grained with 50 types~\cite{li2006learning} like in the TREC QA task\footnote{\url{https://trec.nist.gov/data/qamain.html}}. We propose that a more granular answer type classification is possible and usefule using popular Semantic Web ontologies such as DBepdia and Wikidata.

In addition to answer type prediction task which we also had last year, we will introduce the subtask of relation prediction \cite{rossiello2021generative,naseem2021semantics,earl,mihindukulasooriya2020leveraging,sakor2020falcon,lin2020kbpearl,sakor2019old,pan2019entity} as another step towards a functional KGQA system. This decision is also based on last year's systems performances and their implications. Relation prediction is a vital step towards formulating the correct query to extract the answer from the knowledge base. A question can contain one or more relations from an ontology connecting the entities and classes directly or indirectly mentioned in the question~\cite{DBLP:journals/semweb/UsbeckRHCHNDU19}. Participants will be able to choose to participate in one or both tasks using their systems. The two tasks will maintain two leaderboards.


\section{Task Description}
\subsection{Answer Type Prediction}
Given a natural language question, the task is to produce a ranked list of answer types of a given target ontology. Currently, the target ontology could be either \textit{DBpedia} or \textit{Wikidata}. Table~\ref{tab1:examples} illustrates some examples. The participating systems can be either supervised (training data is provided) or unsupervised.

\begin{table}[htb!]
\centering
\caption{Example questions and answer types}\label{tab1:examples}
\begin{tabular}{|p{7cm}|l|l|}
\hline
\multirow{2}{*}{\textbf{Question}}                   & \multicolumn{2}{c|}{\textbf{Answer Type}}                                      \\ \cline{2-3} 
                                                     & \multicolumn{1}{c|}{\textbf{DBpedia}} & \multicolumn{1}{c|}{\textbf{Wikidata}} \\ \hline
Give me all actors starring in movies directed by and starring William Shatner. & dbo:Actor & wd:Q33999 \\ \hline
Who is the heaviest player of the Chicago Bulls?     & dbo:BasketballPlayer                  & wd:Q3665646                            \\ \hline
How many employees does Google have?     & xsd:integer                  & xsd:integer                             \\ \hline
\end{tabular}
\end{table}

\subsection{Relation Prediction}
The second task is similar to the answer type prediction task. Here, the task is predict a set of relations from a given natural language question based on the underlying knowledge graph. The knowledge graph is either \textit{DBpedia} or \textit{Wikidata}. Table~\ref{tab2:examples} illustrates some examples. The participating systems can be either supervised (training data is provided) or unsupervised.
Note, an answer span detection is not needed nor do implicite relations from general ontologies such as \textit{rdf:type} or \textit{rdfs:label} need to be detected.


\begin{table}[htb!]
\centering
\caption{Example questions and relations}\label{tab2:examples}
\begin{tabular}{|p{6cm}|l|l|}
\hline
\multirow{2}{*}{\textbf{Question}}                   & \multicolumn{2}{c|}{\textbf{Relations}}                                      \\ \cline{2-3} 
                                                     & \multicolumn{1}{c|}{\textbf{DBpedia}} & \multicolumn{1}{c|}{\textbf{Wikidata}} \\ \hline
Give me all actors starring in movies directed by and starring William Shatner. & dbo:starring, dbo:director & wdt:P161, wdt:P57  \\ \hline
Which programming languages were influenced by Perl? & dbo:influencedBy               &              wdt:P737                  \\ \hline
How many employees does Google have?     &  dbo:numberOfEmployees                 &     wdt:P1128                         \\ \hline
\end{tabular}
\end{table}

\section{Datasets}

Rather than building a benchmark from scratch, several datasets for semantic answer type prediction are created from existing academic benchmarks 
for Knowledge Graph Question Answering. For creating the gold standards, we have used QALD-9~\cite{ngomo20189th}, LC-QuAD v1.0~\cite{trivedi2017lc}, and LC-QuAD v2.0~\cite{dubey2019lc} datasets. To overcome the problem of repeating last year's data, we have re-split the datasets and added new datasets. Tentatively, we are planing to add questions from RuBQ~\cite{rubq2}, SimpleQuestionsWikidata~\cite{wikidata-benchmark}, SimpleDBpediaQA~\cite{azmy2018farewell}, WebQSP-WD~\cite{webspwd}, and Complex Sequential Question Answering~\cite{csqa2018} datasets. Additionally, we have added a certain amount of honey pot questions to avoid cheating. 

\textbf{Answer Type Prediction: } 
Each of these datasets has a natural language query and a corresponding SPARQL query. We used the gold standard SPARQL query to generate results and analyze them to generate an initial answer type for each query and manually validated them. Table\ref{tab:ATData}

\begin{table}[h!]
\centering
\caption{Answer Type Prediction Datasets}\label{tab:ATData}
\begin{tabular}{|l|l|l|l|}
\hline
Dataset & Train  & Test   & Total  \\ \hline
SMART2021-AT-DBpedia      & 40,621 & 10,093 & 50,714 \\ \hline
SMART2021-AT-Wikidata       & 43,604 & 10,864 & 54,468 \\ \hline
\end{tabular}
\end{table}

\textbf{Relation Prediction:}
Similar to the answer type prediction task, for the relation prediction task, we extracted the corresponding relations from the given SPARQL queries. The datasets have natural language queries and their corresponding SPARQL queries. Each SPARQL query has a set of triple patterns, each containing a relation. For each natural language questions, those relations are considered as the gold standard relations.

\begin{table}[h!]
\centering
\caption{Relation Prediction Datasets}\label{tab:RLData}
\begin{tabular}{|l|l|l|l|}
\hline
Dataset & Train  & Test   & Total  \\ \hline
SMART2021-RL-DBpedia      & 34,204 & 8,552 & 42,756 \\ \hline
SMART2021-RL-Wikidata       & 24,112 & 6,029 & 30,141 \\ \hline
\end{tabular}
\end{table}


\section{Evaluation metrics and software}

The systems can participate in one or both tasks in one or both datasets. The are expected to provide a list of answer types or/and a set of relations from the target ontology for each natural language question. The answer types will be evaluated using the metrics: \textit{Mean Reciprocal Rank} (\textit{MRR}), \textit{Hits @ 1/5/10}, and \textit{Macro F1}. The systems will be ranked based on F1. The relations will be evaluated using \textit{Precision}, \textit{Recall}, \textit{Micro F1} and  \textit{Macro F1} and ranked according to the latter metric.

We provide evaluation scripts\footnote{\url{https://github.com/smart-task/smart-2021-dataset/tree/main/evaluation}} (in Python) for each task. 

\pagebreak

\bibliographystyle{apalike}  
\bibliography{ms}

\end{document}